\documentclass[sigconf, screen]{acmart}
\AtBeginDocument{%
  }

\setcopyright{acmlicensed}
\copyrightyear{2025}
\acmYear{2025}
\acmDOI{XXXXXXX.XXXXXXX}

\acmConference[MM '25]{ACM International Conference on Multimedia 2025}{October 27--31,
  2025}{Dublin, Ireland}

\acmISBN{978-1-4503-XXXX-X/2018/06}

\definecolor{customcite}{HTML}{e67a7a}
\definecolor{customlink}{HTML}{b83b5e}
\definecolor{customurl}{HTML}{11999e}
\AtEndPreamble{
\usepackage{hyperref}

    \hypersetup{
      colorlinks = true,
      linkcolor = customlink,
      anchorcolor = purple,
      citecolor = customcite,
      filecolor = purple,
      urlcolor = customurl
    }
}

\usepackage{tabularx}  

\begin{document}

\title{MVP: Winning Solution to SMP Challenge 2025 Video Track}

\author{Liliang Ye}
\affiliation{%
  \institution{Huazhong University of Science and Technology}
  \city{Wuhan}
  \country{China}
}
\email{yll@hust.edu.cn}

\author{Yunyao Zhang}
\affiliation{%
  \institution{Huazhong University of Science and Technology}
  \city{Wuhan}
  \country{China}
}
\email{ikoyun@hust.edu.cn}

\author{Yafeng Wu}
\affiliation{%
 \institution{Huazhong University of Science and Technology}
 \city{Wuhan}
 \country{China}
 }
\email{wyf2024@hust.edu.cn}

\author{Yi-Ping Phoebe Chen}
\affiliation{%
 \institution{La Trobe University}
 \city{Melbourne}
 \country{Australia}
 }
\email{phoebe.chen@latrobe.edu.au}

\author{Junqing Yu}
\affiliation{%
 \institution{Huazhong University of Science and Technology}
 \city{Wuhan}
 \country{China}
 }
\email{yjqing@hust.edu.cn}

\author{Wei Yang}
\affiliation{%
 \institution{Huazhong University of Science and Technology}
 \city{Wuhan}
 \country{China}
 }
\email{weiyangcs@hust.edu.cn}

\author{Zikai Song}
\authornote{Corresponding author.}
\affiliation{%
  \institution{Huazhong University of Science and Technology}
  \city{Wuhan}
  \country{China}
}
\email{skyesong@hust.edu.cn}

\renewcommand{\shortauthors}{Liliang Ye et al.}

\begin{abstract}
  Social media platforms serve as central hubs for content dissemination, opinion expression, and public engagement across diverse modalities. Accurately predicting the popularity of social media videos enables valuable applications in content recommendation, trend detection, and audience engagement. In this paper, we present Multimodal Video Predictor (MVP), our winning solution to the Video Track of the SMP Challenge 2025. MVP constructs expressive post representations by integrating deep video features extracted from pretrained models with user metadata and contextual information. The framework applies systematic preprocessing techniques, including log-transformations and outlier removal, to improve model robustness. A gradient-boosted regression model is trained to capture complex patterns across modalities. Our approach ranked first in the official evaluation of the Video Track, demonstrating its effectiveness and reliability for multimodal video popularity prediction on social platforms. The source code is available at \url{https://anonymous.4open.science/r/SMPDVideo}.
\end{abstract}



\maketitle

\section{INTRODUCTION}
\label{introduction}

In recent years, social media platforms have rapidly evolved into dynamic ecosystems for information dissemination and user interaction, with video content emerging as a dominant medium~\cite{ciscoCiscoAnnualInternet2020}. 
The widespread adoption of short videos, live streams, and other multimedia formats has significantly transformed how users express themselves and engage with content~\cite{violot2024shorts}. 
This transformation has led to an unprecedented volume of user-generated videos, posing new challenges and opportunities for understanding content dynamics. 
Among these, accurately predicting the popularity of social media videos has become a critical task, underpinning key applications such as personalized content recommendation, trend detection, and data-driven decision-making for content creators and platforms alike.
However, the popularity of social media videos is influenced by a multitude of factors, including visual features, textual descriptions, user attributes, and contextual information~\cite{meghawatMultimodalApproachPredict2018, xu2020multimodal, wu2024smp}. Effectively integrating multimodal information, constructing expressive feature representations, and leveraging advanced machine learning techniques remain significant challenges in this field.

Recent advances in multimodal learning have integrated visual-language models such as CLIP~\cite{radford2021learning} and VideoMAE~\cite{tongVideoMAEMaskedAutoencoders2022} with structured context features for video popularity prediction. Contemporary approaches typically combine convolutional neural networks for visual feature extraction with transformer-based architectures for textual understanding. However, existing methods face critical limitations: most frameworks treat modalities independently during feature extraction, missing crucial cross-modal dependencies; current visual encoders struggle to capture aesthetic and emotional nuances that drive user engagement; temporal modeling approaches often overlook the rapid evolution of trending topics and viral propagation patterns unique to social platforms.

To further advance the understanding of content dynamics in multimodal environments, the Social Media Prediction (SMP) Challenge 2025 introduces a dedicated Video Track centered on short-form user-generated video analysis. The task is defined as a regression problem: given a video post with visual frames, captions, user information, and posting time, the goal is to predict its popularity score. This score reflects aggregated engagement signals and is evaluated using Mean Absolute Percentage Error (MAPE), which accounts for scale variation in real-world data. The dataset, SMPD-Video, includes 6,000 posts from 4,500 users, along with metadata such as categories, tags, and basic video and user attributes. The challenge emphasizes the complex interplay between content, user behavior, and temporal context. By establishing a standardized benchmark, the SMP Challenge 2025 Video Track fosters progress in multimodal learning for social media applications and offers a rigorous testbed for evaluating video-based popularity prediction systems.

In this study, we propose \textbf{M}ultimodal \textbf{V}ideo \textbf{P}redictor (MVP), a robust framework for social media video popularity prediction. Our method leverages a pretrained XCLIP model to extract deep visual representations from sampled video frames, capturing high-level semantic cues. These features are combined with structured metadata, including user statistics, posting time, and content attributes, which are carefully engineered and log-transformed to improve stability. To mitigate the impact of noise, we apply outlier removal and normalization techniques during preprocessing. A CatBoost~\cite{dorogush2018catboost} regressor is then trained on the fused feature set to model complex cross-modal interactions. Our solution achieves top performance on the SMPD-Video dataset, ranking \textbf{first place} in the Video Track of the SMP Challenge 2025. This result demonstrates the effectiveness of integrating visual content and structured context for accurate popularity prediction.
\section{RELATED WORK}
\label{related-work}

Popularity prediction for visual content originated in image-based analysis~\cite{song2022transformer,song2025temporal,luo2024diffusiontrack}, where models used features such as color composition and caption text to estimate engagement with tree-based regressors like XGBoost~\cite{chen2016xgboost, wu2019smp, wu2023smp, maoEnhancedCatBoostStacking2023, laiHyFeaWinningSolution2020, tuHigherOrderVisionLanguageAlignment2024}. 
These early studies demonstrated the value of combining heterogeneous modalities and inspired extensions to video popularity prediction.

With the advent of video content~\cite{zhou2025video,hu2025sf2t}, researchers introduced multimodal frameworks~\cite{li2024coupled,song2024autogenic,song2021distractor} to address the inherent noise and uncertainty~\cite{song2023compact} present in video data.
For instance, stochastic embeddings combined with a product-of-experts encoder have been utilized to seamlessly integrate video, textual, and metadata information, thereby improving prediction robustness~\cite{zhuPredictingPopularityMicrovideos2020, miech2018learning}.
Other methods have focused on enhancing representation by retrieving similar videos and incorporating their features during inference, which has shown further improvements in predictive accuracy~\cite{liuMultiModalVideoFeature2025a, liu2019use}. Building upon this retrieval paradigm, recent advances have introduced more sophisticated graph-based approaches that exploit similar-content networks to enhance post representations~\cite{cheng2024retrieval}. These retrieval-augmented modeling techniques leverage external knowledge bases to enrich feature representations with contextual information, while hypergraph-based methods capture complex multi-relational dependencies between users, content, and engagement patterns.
The extraction of deep frame-level features from state-of-the-art pretrained video backbones, such as TimeSformer~\cite{bertasiusSpaceTimeAttentionAll2021}, ViViT~\cite{arnabViViTVideoVision2021}, VideoMAE~\cite{tongVideoMAEMaskedAutoencoders2022}, and X-CLIP~\cite{maXCLIPEndtoEndMultigrained2022}, has become increasingly prevalent.
A representative approach combines such embeddings with BERT-encoded captions and structured metadata, followed by ensemble learning through neural networks and XGBoost to enhance predictive performance~\cite{liuMultiModalVideoFeature2025a}. 
Concurrently, researchers have explored user bias disentanglement techniques to improve model generalization across diverse user populations~\cite{hu2024dual}, addressing the challenge of personalized engagement patterns that vary significantly across different user demographics and behavioral clusters.
At the same time, vision‑language methods transform video frames and captions into descriptive text tokens, using language models for interpretable engagement estimation~\cite{nguyenVideoLanguageUnderstandingSurvey2024, wang2022language, sun2019videobert}.
Despite these advances, several challenges persist. The highly skewed distribution of video popularity, the presence of noisy or incomplete metadata, and the dynamic nature of user engagement complicate model training and evaluation. The growing importance of structural adaptation and social context modeling in popularity prediction reflects the need for more sophisticated approaches that can handle the complex interplay between content characteristics and social dynamics.

Across these diverse methodologies, several common threads emerge: the integration of multimodal features, the adoption of powerful pretrained representation backbones, and the application of tree-based or hybrid ensemble models. These insights have directly informed the design and development of our MVP framework, which seeks to unify these best practices for robust and interpretable video popularity prediction.

\section{METHODOLOGY}
\label{methodology}

\begin{figure*}[h!]
    \centering
    \includegraphics[width=\linewidth]{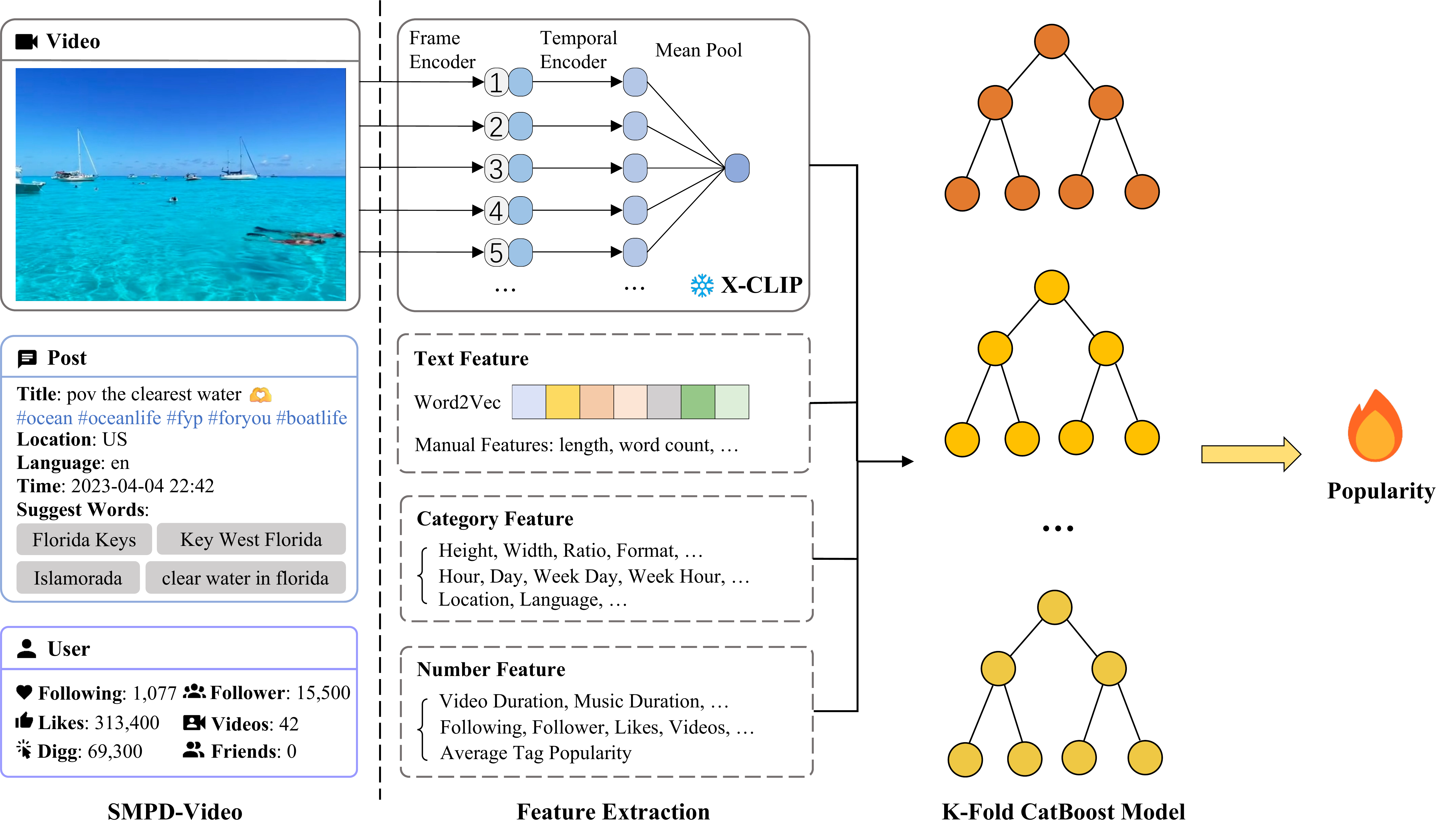} 
    \caption{The MVP framework pipeline processes semantic visual features, user profiles, temporal signals, and metadata for multimodal prediction of video popularity in social media using a gradient-boosted regression model.}
    \label{fig:framework}
\end{figure*}

\subsection{Overview}
In this section, we present the design of the Multimodal Video Predictor (MVP) framework, which systematically addresses the challenge of predicting video popularity in social media contexts. 
As illustrated in ~\autoref{fig:framework}, the architecture of MVP is tailored to integrate heterogeneous sources of information and model complex interactions across multiple modalities. 
The pipeline begins with the extraction of high-level semantic representations from video content using a pretrained visual encoder. 
These visual embeddings are then concatenated with structured features derived from user profiles, temporal signals, and post metadata. 
To ensure feature compatibility and minimize the influence of noise, we implement a series of preprocessing steps, including log transformation, normalization, and outlier removal. 
The resulting multimodal feature set is subsequently used to train a gradient-boosted regression model, which is capable of capturing intricate cross-modal dependencies for accurate popularity estimation. 
The overall framework is designed to be robust and interpretable, facilitating reliable modeling of the diverse and dynamic nature of social media video data. 
In the following subsections, we describe each component of the pipeline in detail.

\subsection{Multimodal Feature Construction and Fusion}

A central aspect of the MVP framework is the systematic extraction and integration of multimodal features that capture the diverse factors influencing social media video popularity. We organize feature engineering into three principal components: visual features, user-related features, and temporal-metadata features, each offering distinct perspectives on user-generated content.

\textbf{Visual Feature Extraction.}
We employ a pretrained XCLIP encoder to extract deep visual representations from each video. For every video, we uniformly sample a fixed number of frames to ensure temporal coverage. We process each frame with the XCLIP model, generating high-dimensional embeddings that represent semantic cues such as scene context, objects, and actions. After obtaining these frame-level embeddings, we apply average pooling to aggregate information across frames, resulting in a compact video-level feature. To further balance expressiveness and computational efficiency, we use Principal Component Analysis (PCA)~\cite{abdi2010principal} to reduce the dimensionality of the stacked embeddings.

Specifically, let $\mathbf{v}_i \in \mathbb{R}^d$ denote the embedding of the $i$-th sampled frame, and $N$ be the total number of frames. The video-level feature $\mathbf{v}_{video}$ is computed by average pooling:
\begin{equation}
\mathbf{v}_{video} = \mathbf{W}^T \left( \frac{1}{N} \sum_{i=1}^N \mathbf{v}_i - \boldsymbol{\mu} \right)
\end{equation}
where $\mathbf{W}$ is the PCA projection matrix and $\boldsymbol{\mu}$ is the mean vector. This process yields a compact and expressive feature for each video, facilitating robust downstream modeling.

\textbf{User-Related Features.}
To characterize the influence of users, we extract detailed user profile statistics. These features include follower count, following count, total published videos, cumulative likes, diggs, hearts, friends, and the average popularity of historical posts for each user. We apply logarithmic transformation to all user-related counts to reduce skewness and enhance numerical stability. These features capture both the social reach and historical activity of the user, which are essential for understanding content dissemination.

\textbf{Temporal, Metadata, and Semantic Features.}
To capture periodic engagement patterns, we encode posting time using multiple categorical features, such as hour of day and day of week. Each post's metadata includes video category, tags, language, location, and video-specific attributes like resolution, aspect ratio, duration, and music identifiers. Further feature enrichment involves processing video captions and suggested keywords with standard natural language processing steps, including lowercasing, tokenization, and embedding extraction using a pretrained Word2Vec model. This process yields additional features, such as caption length, token count, and the number of suggested keywords. Additionally, computing the mean popularity of tags for each post provides information about topic trends and audience preferences. Finally, label encoding or embedding for categorical metadata and normalization of all continuous features ensure consistency across modalities.

\textbf{Feature Fusion and Preprocessing.}
We concatenate all extracted features to form a unified multimodal feature vector for each post. Formally, let $\mathbf{v}_{visual}$, $\mathbf{v}_{user}$, and $\mathbf{v}_{meta}$ represent the visual, user-related, and temporal-metadata feature vectors, respectively. The final multimodal feature vector $\mathbf{v}_{multi}$ is constructed as:
\begin{equation}
\mathbf{x} = \mathbf{v}_{multi} = [\mathbf{v}_{visual};\mathbf{v}_{user};\mathbf{v}_{meta}]
\end{equation}
where $[\cdot;\cdot;\cdot]$ denotes the concatenation operation. Before feeding the features into the model, we impute missing values with context-appropriate defaults, filter outliers using the interquartile range, and normalize continuous variables. To ensure reliable regression, we preprocess the training labels by removing outliers outside the $[Q_1 - 1.5 \times IQR, Q_3 + 1.5 \times IQR]$ range, where $Q_1$ and $Q_3$ denote the first and third quartiles, and $IQR$ is the interquartile range. This approach ensures stable feature distributions and minimizes the impact of noisy or extreme values, supporting robust model training.
By combining deep semantic visual descriptors, comprehensive user and contextual metadata, temporal signals, and rich textual features, we create a unified and expressive representation for downstream regression modeling. This integrated approach enables the MVP framework to effectively capture the multifaceted drivers of video popularity in social media environments.

\subsection{Regression Model}
To estimate the popularity score of each video post, we employ CatBoost as the regression model. CatBoost is a gradient boosting decision tree method specifically designed for structured data with numerous categorical features. In our framework, all engineered features, including visual descriptors, user attributes, metadata, and temporal signals, are concatenated to form a comprehensive input vector $\mathbf{x}_i$ for each post $i$.

During training, CatBoost utilizes category-aware encoding strategies to transform categorical features such as user identifiers, video format, music title, and posting time. The regression objective is formulated as the minimization of the Huber loss function, which can be written as:
\begin{equation}
\mathcal{L}_{\delta}(y, \hat{y}) =
\begin{cases}
\frac{1}{2}(y - \hat{y})^2, & \text{if } \left| y - \hat{y} \right| \leq \delta \\
\delta \left| y - \hat{y} \right| - \frac{1}{2} \delta^2, & \text{otherwise}
\end{cases}
\end{equation}
where $y$ denotes the ground-truth popularity score and $\hat{y}$ is the predicted value. The parameter $\delta$ determines the transition point between quadratic and linear behavior, thereby improving robustness to outliers while maintaining sensitivity to regular samples.

We apply five-fold cross-validation by splitting the training set into five subsets and iteratively using four for training and one for validation. For each fold $k$, the CatBoost regressor $f^{(k)}$ is trained to learn the mapping:
\begin{equation}
\hat{y}_i^{(k)} = f^{(k)}(\mathbf{x}_i)
\end{equation}
The final prediction $\hat{y}_i$ is calculated by averaging the outputs of all folds:
\begin{equation}
\hat{y}_i = \frac{1}{K} \sum_{k=1}^{K} \hat{y}_i^{(k)}
\end{equation}
where $K=5$ in our experiments.

\section{EXPERIMENT}
\label{experiment}

\subsection{Dataset}
The SMPD-Video dataset comprises 6,000 video posts from 4,500 users, spanning 24 months and covering 120 categories. Each post is annotated with detailed metadata, including user statistics, video attributes, posting time, location, captions, and over 40,000 unique tags. The dataset supports multimodal video popularity prediction with comprehensive visual, textual, and user-related information.

To address the substantial variance in engagement metrics across different videos, where view counts can range from zero to millions, the dataset employs a logarithmic transformation to standardize the popularity labels. Following established practices~\cite{wu2016unfolding} in social media analysis, the normalized popularity score is computed as:
\begin{equation}
s = \log_2 \frac{r}{d} + 1
\end{equation}
where $s$ represents the transformed popularity score, $r$ denotes the raw view count, and $d$ indicates the number of days elapsed since publication. This normalization approach effectively reduces the impact of extreme values while preserving the relative ranking of content popularity, thereby facilitating more stable model training and evaluation.

\begin{figure}[h!]
    \centering
    \includegraphics[width=1\linewidth]{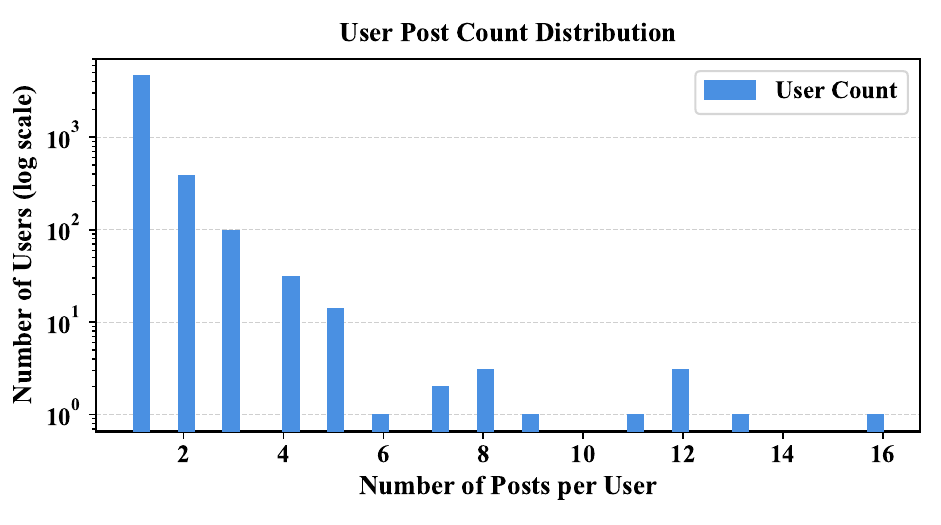}
    \caption{Distribution of user post counts in the SMPD-Video dataset.}
    \label{fig:user_post_distribution}
\end{figure}

~\autoref{fig:user_post_distribution} presents the distribution of user post counts in the SMPD-Video dataset. Most users contribute only a small number of posts, while a minority generate many posts. This long-tailed pattern reflects typical user activity on social media platforms. In comparison with SMPD-Image~\cite{wu2023smp}, the SMPD-Video dataset covers fewer posts per user, as the collection process yields a more limited number of videos for each user. As a result, the prediction task in the video track places greater emphasis on the features of individual videos and their content, rather than relying on extensive user history.

\subsection{Evaluation Metrics}

The effectiveness of our approach is evaluated using the Mean Absolute Percentage Error (MAPE). As a scale-invariant metric, MAPE is suitable for measuring prediction accuracy across diverse popularity values. For a set of $n$ samples, let $y_i$ denote the ground-truth popularity and $\hat{y}_i$ the predicted value for the $i$-th video. The MAPE is defined as
\begin{equation}
    \mathrm{MAPE} = \frac{1}{n} \sum_{i=1}^{n} \left| \frac{y_i - \hat{y}_i}{y_i} \right|
\end{equation}
MAPE provides a scale-invariant measure of accuracy, making it particularly suitable for popularity prediction tasks with large value ranges. 

\subsection{Main Results}

\subsubsection{Overall Performance}

To assess the effectiveness of the proposed approach, we report the final performance on the official evaluation set using the Mean Absolute Percentage Error (MAPE) as the primary metric. Our solution achieves a MAPE of \textbf{0.1754}. This result demonstrates the superiority of the designed model architecture and feature engineering pipeline in capturing the intricate relationships underlying social media popularity prediction.

\subsubsection{Ablation Study}

We conduct a comprehensive ablation study to clarify the contribution of each model component and design choice. ~\autoref{tab:ablation} summarizes the results as we systematically remove or alter each major module and feature group. The full model achieves a MAPE of 0.1754, establishing a strong performance baseline. Removing video embedding features increases the MAPE to 0.1818, which demonstrates the crucial role of visual information for popularity prediction. Without text embedding features, the MAPE rises to 0.1810, confirming that semantic information from textual content supports accurate predictions. When we exclude tag popularity statistics, the MAPE reaches 0.1785, indicating that social trend signals provide valuable context for the model. Removing video metadata features results in a MAPE of 0.1782, while eliminating temporal features leads to a MAPE of 0.1758. These results show that both technical attributes and temporal signals offer additional, though moderate, improvements in prediction accuracy.

We also investigate the impact of preprocessing and model aggregation strategies. When we remove the outlier removal procedure, the MAPE increases to 0.1839, which highlights the necessity of robust data cleaning to handle noisy social media inputs. Eliminating the K-fold ensemble strategy results in a MAPE of 0.1786, underscoring the performance gains achieved through model aggregation. Among all feature groups, user-related features prove most influential. Excluding these features causes the MAPE to surge to 0.3010, revealing that user engagement history and social network characteristics represent primary predictors for content virality. This dramatic performance drop illustrates how information about user behavior and connections enables the model to capture underlying popularity dynamics that other features cannot provide.

Overall, the ablation results demonstrate that the model's success arises from the interplay of multimodal feature extraction, effective data preprocessing, and ensemble learning. Each component contributes to performance improvements, while user-centric and semantic features deliver especially significant benefits. These findings highlight the importance of designing a holistic approach that integrates diverse sources of information and leverages advanced data processing techniques to maximize predictive accuracy.

\begin{table}[!htbp]
    \centering
    \caption{Ablation study results on the validation set.}
    \begin{tabularx}{0.8\linewidth}{X c}
        \toprule[1.2pt]
        \textbf{Methods} & \textbf{MAPE} \\
        \midrule
        MVP & \textbf{0.1754} \\
        w/o Video Embedding & 0.1818 \\
        w/o Text Embedding & 0.1810 \\
        w/o Filter & 0.1839 \\
        w/o User Profile & 0.3010 \\
        w/o Video Metadata & 0.1782 \\
        w/o Tag Popularity & 0.1785 \\
        w/o Time & 0.1758 \\
        w/o K-Fold & 0.1786 \\
        \bottomrule[1.2pt]
    \end{tabularx}
    \label{tab:ablation}
\end{table}

\subsubsection{Feature Importance}
To understand which features contribute most to the model’s prediction, we analyze the feature importance scores output by CatBoost. ~\autoref{tab:importance} lists the top 20 features ranked by their importance.

Among these features, user engagement indicators such as \textit{LikeCount}, \textit{VideoCount}, and \textit{HeartCount} are assigned the highest importance, suggesting that historical activity and user popularity play a substantial role in video popularity prediction. Additionally, several user-related and content-based features, including \textit{FollowerCount} and \textit{SuggestedWordsLen}, demonstrate considerable influence on the model’s output. It is noteworthy that latent features extracted by dimensionality reduction methods, such as the \textit{svd\_mode} series and \textit{video\_embedding} vectors, also appear frequently in the top ranks, indicating that the model benefits from both explicit and implicit representations of multimodal data.

The distribution of importance scores reflects the complementary effect of combining user metadata, textual information, and deep video embeddings within a unified framework. These findings confirm that a balanced integration of heterogeneous features can enhance predictive performance for the popularity estimation task. Furthermore, the relative importance of user-related features highlights the continued influence of user history and social reach in content dissemination on social platforms.

\begin{table}[!htbp]
    \centering
    \caption{Importance of different features.}
    \resizebox{1\linewidth}{!}{%
    \begin{tabular}{llc|llc}
        \toprule[1.2pt]
        \textbf{Rank} & \textbf{Feature} & \textbf{Importance} & \textbf{Rank} & \textbf{Feature} & \textbf{Importance} \\
        \midrule
        1  & LikeCount         & 13.775 & 11 & FollowingCount  & 1.739 \\
        2  & VideoCount         & 12.392 & 12 & svd\_mode\_t\_11        & 1.137 \\
        3  & HeartCount         & 12.033 & 13 & svd\_mode\_t\_16        & 1.130 \\
        4  & FollowerCount      & 9.986  & 14 & svd\_mode\_t\_1         & 1.113 \\
        5  & SuggestedWordsLen & 4.184  & 15 & svd\_mode\_4            & 1.009 \\
        6  & AvgTagPopularity       & 3.822  & 16 & svd\_mode\_7            & 0.945 \\
        7  & svd\_mode\_0               & 2.621  & 17 & PostContentLen      & 0.942 \\
        8  & PostCount                 & 2.529  & 18 & video\_embedding\_15    & 0.887 \\
        9  & PostLocation             & 2.492  & 19 & svd\_mode\_t\_18        & 0.868 \\
        10 & svd\_mode\_t\_2            & 2.058  & 20 & video\_embedding\_19    & 0.820 \\
        \bottomrule[1.2pt]
    \end{tabular}}
    \label{tab:importance}
\end{table}

\subsection{Analysis}
\begin{figure}[h!]
    \centering
    \includegraphics[width=1\linewidth]{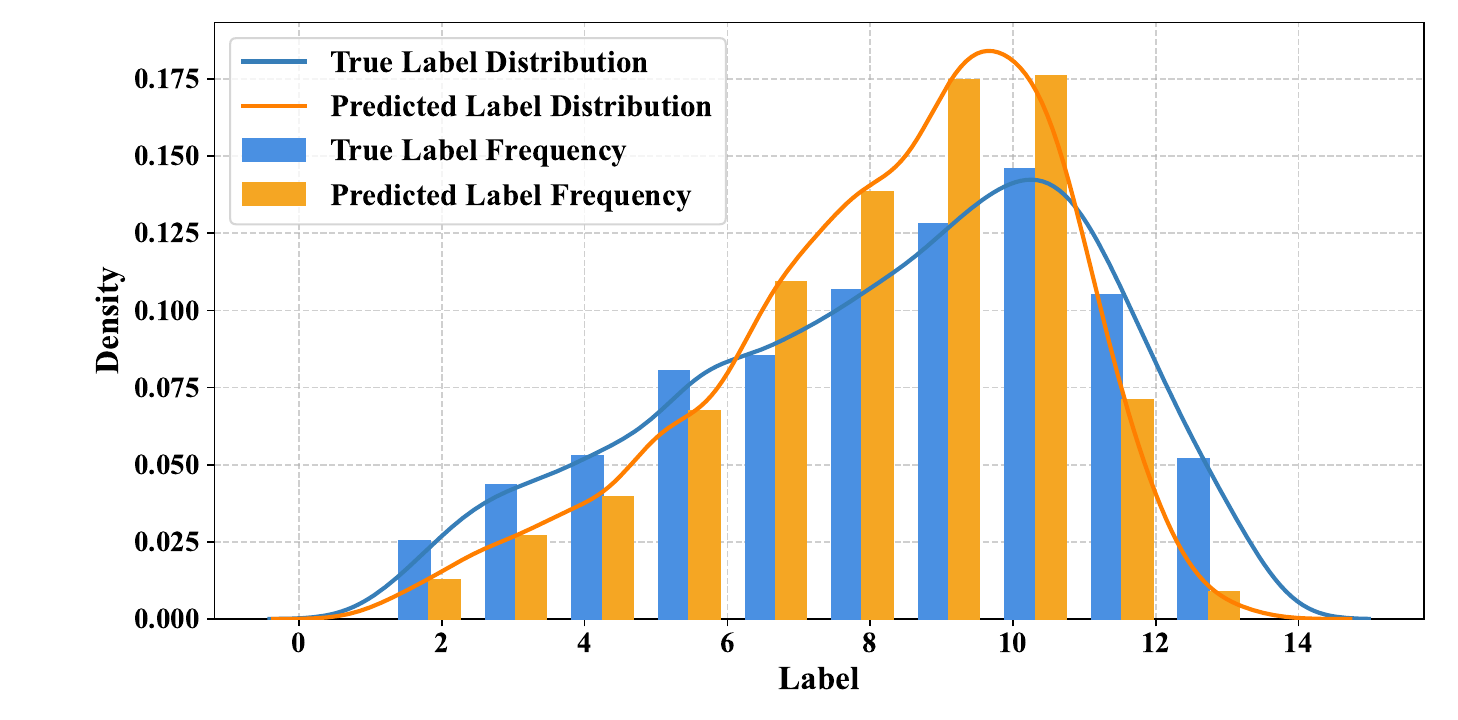}
    \caption{Histogram and kernel density estimation of predicted vs. ground-truth label distributions on the validation set.}
    \label{fig:label_density}
\end{figure}

To further explore the alignment between our model's predictions and the true popularity scores, we conduct a statistical analysis on the validation set. Figure~\autoref{fig:label_density} illustrates the histograms and density distributions for both predicted and actual labels.

The predicted scores demonstrate a strong correspondence to the empirical distribution of the ground-truth labels, effectively capturing the unimodal and skewed nature of the data. Notably, the peak density of the predictions occurs in a similar range as the true scores (approximately 5 to 10), indicating that the dominant pattern present in the data is accurately learned by the model. A slight smoothing in the prediction curve is observed, which is likely a result of ensemble inference and regularization strategies designed to mitigate overfitting.

Despite this overall alignment, the model shows relatively conservative estimates at the distribution extremes. While predictions span the entire label range, there is a tendency to underestimate scores in the low-popularity (<2.5) and high-popularity (>15) intervals. This phenomenon is common in regression tasks involving long-tailed distributions and can be partially attributed to the imbalance of training samples in these regions, as reflected in the ground-truth histogram. To address this limitation, future work may consider techniques such as label distribution smoothing or cost-sensitive re-weighting to improve model performance on rare cases.

Overall, the predicted results exhibit a high degree of statistical fidelity to the actual label distribution. The model is capable of producing scores with meaningful variance and reliable calibration, supporting its potential for large-scale popularity prediction in realistic social media scenarios.

\section{CONCLUSION AND FUTURE WORK}
\label{conclusion-and-future-work}

This work presents a unified and robust framework for predicting the popularity of social media videos, leveraging comprehensive multimodal feature extraction, advanced machine learning models, and ensemble strategies. Through careful integration of visual, textual, user, and structural information, the proposed approach effectively captures the complex patterns underlying social engagement dynamics. Experimental results demonstrate that incorporating domain-specific features and explicitly modeling user behavior can substantially improve predictive accuracy.

The MVP framework exhibits strong potential for cross-platform transferability, as the core multimodal feature extraction and ensemble learning components can be adapted to different social media environments with minimal architectural modifications. The modular design facilitates deployment across various content types and user demographics, making it suitable for real-world applications in content recommendation systems and marketing analytics.

Nonetheless, challenges such as semantic inconsistencies across modalities and the dynamic nature of social media content remain open for further exploration~\cite{baltruvsaitis2018multimodal}. Future research may focus on developing more adaptive multimodal fusion techniques, as well as end-to-end architectures capable of leveraging temporal patterns and contextual cues. Moreover, enhancing the interpretability and robustness of prediction systems through improved alignment and representation learning is a promising direction.

In future work, the MVP framework can be extended to support online popularity forecasting under streaming settings, or adapted to incorporate graph-based user interaction structures and multi-task objectives, such as virality classification and trend forecasting. Graph neural networks could capture complex social network dynamics and user influence propagation patterns, while multi-task learning approaches might simultaneously predict engagement metrics, content lifespan, and audience demographics. Additionally, developing real-time deployment strategies for continuous model updating and adaptive threshold adjustment would enhance practical applicability in dynamic social media environments.

In summary, this study provides a meaningful foundation for further multimodal social media analysis and offers valuable insights into building more accurate and generalizable content popularity prediction models.


\bibliographystyle{ACM-Reference-Format}
\bibliography{main}

\end{document}